\documentclass[letterpaper, 10 pt, conference]{ieeeconf}  

\IEEEoverridecommandlockouts                              

\usepackage{graphics} 
\usepackage{algorithm}
\usepackage[noend]{algpseudocode}

\usepackage{amsmath} 
\usepackage{amssymb}  
\usepackage{gensymb}

\usepackage{graphicx}
\usepackage{booktabs}
\usepackage{duckuments}
\usepackage{subfiles} 
\usepackage{acro}

\usepackage{caption}
\usepackage{subcaption}

\usepackage[pagebackref,breaklinks,colorlinks]{hyperref}

\DeclareAcronym{paper_name}{
    short = PEnG,
    long  = Pose-Enhanced Geo-Localisation,
    tag   = paper
}
\DeclareAcronym{cvgl}{
    short = CVGL,
    long  = Cross-View Geo-localisation,
    tag   = nomencl
}
\DeclareAcronym{fov}{
    short = FOV,
    long  = Field-of-View,
    tag   = nomencl
}
\DeclareAcronym{fovs}{
    short = FOVs,
    long  = Fields-of-View,
    tag   = nomencl
}
\DeclareAcronym{bev}{
    short = BEV,
    long  = Birds-Eye-View,
    tag   = nomencl
}
\DeclareAcronym{gnss}{
    short = GNSS,
    long  = Global Navigation Satellite Systems,
    tag   = nomencl
}
\DeclareAcronym{sota}{
    short = SOTA,
    long  = state of the art,
    tag   = nomencl
}
\DeclareAcronym{ape}{
    short = APE,
    long  = Absolute Pose Estimation,
    tag   = nomencl
}
\DeclareAcronym{rpe}{
    short = RPE,
    long  = Relative Pose Estimation,
    tag   = nomencl
}
\DeclareAcronym{dof}{
    short = DoF,
    long  = Degrees of Freedom,
    tag   = nomencl
}
\DeclareAcronym{cdf}{
    short = CDF,
    long  = Cumulative Distribution Function,
    tag   = nomencl
}

\newcommand{\medninety}{22.77}
\newcommand{\toponeninety}{9.12}
\newcommand{\topfiveninety}{29.18}
\newcommand{\toptwentyfiveninety}{51.37}
\newcommand{\topfiveincrease}{213\%}
\newcommand{\meddecrease}{96.90\%}

\title{\LARGE \bf
PEnG: Pose-Enhanced Geo-Localisation
}

\author{Tavis Shore$^{1}$ and Oscar Mendez$^{1}$ and Simon Hadfield$^{1}$
\thanks{$^{1}$ Centre for Vision Speech and Signal Processing, University of Surrey,
Guildford, United Kingdom, \{\tt\small t.shore, o.mendez, s.hadfield\}@surrey.ac.uk}%
}

\begin{document}
\maketitle
\thispagestyle{empty}
\pagestyle{empty}

\begin{abstract}
Cross-view Geo-localisation is typically performed at a coarse granularity, because densely sampled satellite image patches overlap heavily.
This heavy overlap would make disambiguating patches very challenging.
However, by opting for sparsely sampled patches, prior work has placed an artificial upper bound on the localisation accuracy that is possible.
Even a perfect oracle system cannot achieve accuracy greater than the average separation of the tiles. 
To solve this limitation, we propose combining cross-view geo-localisation and relative pose estimation to increase precision to a level practical for real-world application. 
We develop \acs{paper_name}, a 2-stage system which first predicts the most likely edges from a city-scale graph representation upon which a query image lies. 
It then performs relative pose estimation within these edges to determine a precise position.
\acs{paper_name} presents the first technique to utilise both viewpoints available within cross-view geo-localisation datasets to enhance precision to a sub-metre level, with some examples achieving centimetre level accuracy.
Our proposed ensemble achieves state-of-the-art precision - with relative Top-5m retrieval improvements on previous works of \topfiveincrease{}. Decreasing the median euclidean distance error by \meddecrease{} from the previous best of 734m down to \medninety{}m, when evaluating with $\textbf{90\degree}$ horizontal FOV images. 
Code will be made available: \href{https://tavisshore.co.uk/peng}{tavisshore.co.uk/PEnG}.\\


\vspace{0.1em}

\end{abstract}

\section{Introduction}
Localisation is vital in the majority of mobile robotics applications. 
Common techniques such as \ac{gnss} provide absolute positioning data to clients. 
These are prone to failure in certain environments. One example are dense urban canyons such as New York City where tall buildings cause signal occlusions \& reflections, preventing successful satellite communication.
Another example are regions of conflict where malicious actors purposefully  disrupt positioning by spoofing signals, inserting erroneous information. 

\vspace{0.4em}

Image localisation may provide a solution as agents can fully self-localise using onboard sensors, removing requirements for external communication. 
These techniques aim to relate an agent's query image with previously seen geo-tagged images, determining an updated position according to feature and positional similarities with these references. 
A large proportion of mobile robots are already equipped with cameras, increasing the viability of image localisation.

\ac{cvgl} is an increasingly popular branch of image localisation research, offering a viable form of generalisable wide-scale image localisation.
The objective is to relate a street-level query image to a database of reference satellite images - returning the geographic coordinates of the highest correlating known satellite image.

\vspace{0.4em}

Pose estimation is a related field aiming to determine a camera's pose within a scene.
These techniques generally operate at a smaller scale than \ac{cvgl}, localising within a few metres, instead of whole cities. 
They generally operate as continuous prediction, rather than retrieval problems, and operate in N-\ac{dof} as opposed to simple geographic coordinates.
Pose estimation has two primary sub-fields - \ac{ape} and \ac{rpe}. 
\ac{ape} aims to determine a camera's position and orientation within a 3D world coordinate frame. \ac{rpe} aims to compute the same, but with respect to a reference camera. 

\begin{figure}[t!]
    \centering
    \includegraphics[width=\columnwidth]{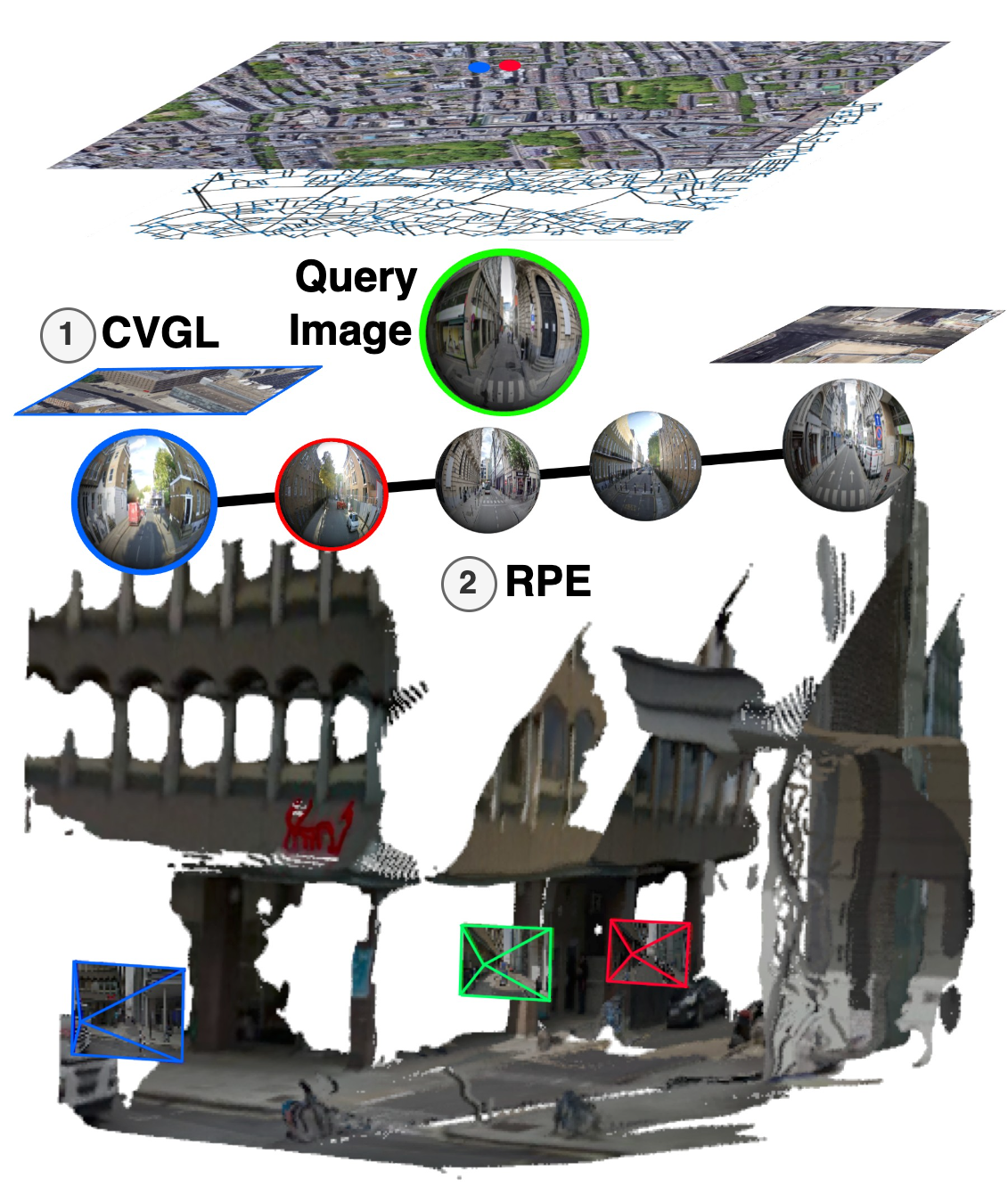}
    \caption{
    \ac{paper_name} Stages: 1) City-scale satellite image with underlying graph network, CVGL estimates candidate edges within city's graph.
    2) Pose estimation along these edges achieves refined geographic poses.
    Green denotes a query input, blue and red display two known reference images.
    }
    \label{fig:frontimage}
    \vspace{-1.25em}
\end{figure}

\vspace{0.4em}

We propose leveraging the advantages of both techniques in a single two-stage system to achieve high-precision city-scale localisation, shown in top-down order in Figure \ref{fig:frontimage}.
Taking as input a street-level image - the first stage performs city-wide \ac{cvgl}, predicting the most recently observed road junction. 
Operating the \ac{cvgl} stage at the scale of road junctions, helps to keep the reference set lean and discriminative, ensuring efficient and accurate retrieval results of coarse location.
The second stage takes the \ac{cvgl} sub-region predictions and performs \ac{rpe} along neighbouring roads, merging likelihoods from both stages to determine a final 3-\ac{dof} pose. 
This novel combination of learned computer vision techniques achieves a reduction in the median localisation error from 734m to \medninety{}m, evaluating with $90\degree$ crops of the StreetLearn dataset \cite{streetlearn}. 

\vspace{0.4em}

\noindent
In summary, our research contributions are:

\begin{itemize}
    \item Introduce the first technique for performing precise image localisation in a city-scale by utilising information from both image viewpoints in \ac{cvgl} datasets. 
    \item Introduce emulating a simple compass, filtering reference embeddings according to a configurable yaw threshold, greatly increasing localisation precision.
    \item Demonstrate strong generalisation to cities not seen in training - localising with a median error of \medninety{}m within the large dense region of Manhattan, considering a region area of $36.1km^2$.
\end{itemize}

\section{Related Works}

\subsection{Camera Pose Estimation}
\ac{rpe} can be divided into two categories: feature matching, and pose regression.
More traditional camera localisation techniques often utilise structure-based methods, representing a scene with an explicit SfM or SLAM reconstruction \cite{old_sfm1, 5206587, 7572201}.
This often requires a large number of images to have already been captured within a scene, limiting generalisation. 

\vspace{0.4em}

Shotton et al. \cite{Shotton2013SceneCR} introduce a novel method called Scene Coordinate Regression Forest (SCoRe Forest) for inferring the pose of an RGB-D camera relative to a known 3D scene using a single image with decision forests.
Kendall et al. propose PoseNet \cite{kendall2015posenet}, the first CNN designed for end-to-end 6-DOF camera pose localisation, evaluating the network thoroughly to prove the viability of deep learning for the field. 
In their following paper \cite{kendall2017geometriclossfunctionscamera}, they apply a principled loss function based on the scene’s geometry to learn camera pose without any hyper-parameters, achieving \ac{sota} results, reducing the performance gap to traditional methods.
Sattler et al. \cite{7572201} propose using a prioritised matching approach, considering features more likely to yield 2D-to-3D matches, terminating searches once sufficient matches have been found.
Brachmann et al. \cite{brachmann2018dsacdifferentiableransac} propose \textit{DSAC}, a differentiable counterpart to RANSAC, replacing the deterministic hypothesis selection with a probabilistic selection, deriving the expected loss with respect to all learnable parameters.
Applying this to image localisation achieved higher accuracies than previous deep learning based methods.
Clark et al. \cite{clark2017vidlocdeepspatiotemporalmodel} propose extending to sequential camera pose estimation, designing an RNN which achieves smoothed poses and greatly reduced localisation error.
Sarlin et al \cite{sarlin2019coarse} propose \textit{HFNet} - performing coarse-to-fine image localisation by predicting local features and global descriptors for 6-DoF localisation simultaneously.
Map-free Relocalisation \cite{arnold2022mapfree} introduces using a single photo from a scene for metric scaled re-localisation, negating the  requirement to construct a scaled map of the scene. 
Rockwell et al. \cite{Rockwell2024} propose \textit{FAR}, combining correspondence estimation and pose regression techniques to utilise the benefits from both to provide  precision and generalisation. 
Wang et al. \cite{wang2023dust3rgeometric3dvision} and Leroy et al. in the follow-up paper \cite{leroy2024groundingimagematching3d} propose \textit{Dust3r} and \textit{Mast3r} respectively. 
Both are techniques for dense unconstrained stereo
3D reconstruction of arbitrary image collections, with no prior information. 
Mast3r achieves \ac{sota} performance in various fields including camera calibration and dense 3D reconstruction.
Moreau1 et al. \cite{moreau2023crossfire} propose \textit{CROSSFIRE} - using NeRFs as implicit scene maps and propose a camera re-localisation algorithm for this representation. 
\textit{CROSSFIRE} achieves \ac{sota} accuracy and is capable of operating in dynamic outdoor environments. 

\vspace{0.4em}

Similar to how \textit{FAR} proposed combining multiple pose estimation paradigms to achieve \ac{sota} performance in that particular sub-field, we propose combining multiple image localisation techniques to achieve high precision localisation in large scale regions with different input modalities.

\subsection{Cross-View Geo-Localisation}
Current \ac{cvgl} techniques primarily focus on embedding retrieval - extracting reduced dimensionality representations of reference satellite images, aiming to return geo-coordinates from those most similar to query images.
Techniques are being increasingly proposed to improve performance by manipulating extracted features, \cite{Zhu2022TransGeoTI}, \cite{SAIG}, \cite{shore2023bevcv}. 

\vspace{0.4em}

Workman and Jacobs \cite{7301385} first propose CNNs for learning feature relationships across viewpoints. This was extended by Lin et al. \cite{7299135}, treating each query uniquely, utilising euclidean similarities for retrieval.
Vo and Hays \cite{Vo2016LocalizingAO} add rotation information through an auxiliary loss, evaluating misalignment impact. 
CVM-Net \cite{8578856} add NetVLAD \cite{netvlad} to the CNN, aggregating local feature residuals to cluster centroids.
Liu and Li \cite{8954224} increase access to orientation information, improving the latent space robustness. 
Shi et al. \cite{Shi2019SpatialAwareFA} developed a spatial attention mechanism, improving feature alignment between views. 
In \cite{Shi2019OptimalFT} they increase the cross-view feature similarity, by applying the techniques to limited-\ac{fov} data. This was important due to the ubiquity of monocular cameras compared with panoramic cameras, increasing feasibility.
\cite{Shi2020WhereAI} computes feature correlation between ground-level images and polar-transformed aerial images, shifting and cropping at the strongest alignment before performing image retrieval. 
Toker et al. \cite{Toker2021ComingDT} synthesised streetview images from aerial image queries before performing image retrieval. 
L2LTR \cite{Yang2021CrossviewGW} developed a CNN+Transformer network, combining a ResNet backbone with a vanilla ViT encoder to increase performance over \ac{sota}.
TransGeo \cite{Zhu2022TransGeoTI} proposed a transformer that uses an attention-guided non-uniform cropping strategy to remove uninformative areas.

In GeoDTR \cite{GeoDTR, GeoDTR+}, Zhang et al. separate geometric information from the raw features, learning spatial correlations within visual features to enhance performance.
Zhu et al. introduced \textit{SAIG} \cite{SAIG}, an attention-based \ac{cvgl} backbone, representing long-range interactions among patches and cross-view associations with multi-head self-attention layers.
BEV-CV \cite{shore2023bevcv} introduces \ac{bev} transforms to the field, reducing representational differences between viewpoints to create more similar embeddings. 
Sample4Geo \cite{sample4geo} propose two \ac{cvgl} sampling strategies, geographically sampling for optimal training initialisation, mining hard-negatives according to feature similarities between viewpoints. 
SpaGBOL \cite{spagbol} propose progressing the CVGL field from single and sequential representations to graph-based representation, allowing for more geo-spatially strong embeddings. 

\begin{figure}[t!]
  \centering
    \includegraphics[width=0.9\columnwidth]{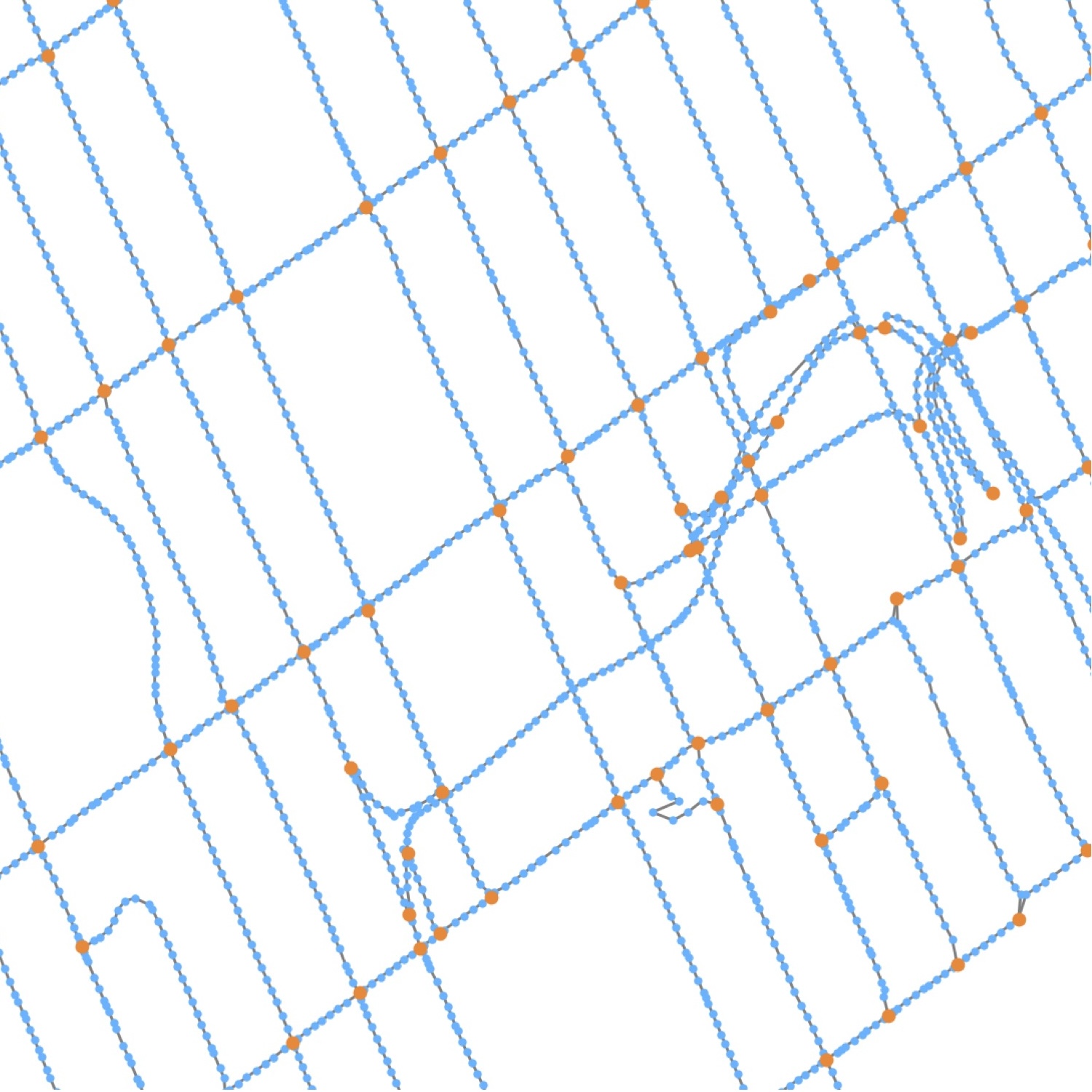}
    \caption{Section of Manhattan graph with primary (orange) and secondary (blue) nodes displayed. Most edges have a constant yaw, motivating the utilisation of a compass.}
    \label{fig:london_graph}
\end{figure}

\vspace{0.4em}

To date all of the above \ac{cvgl} approaches have followed a retrieval paradigm where the accuracy of results is limited by the granularity of the geo-referenced database. 
Sparsely sampled data can lead to higher retrieval rates due to greater feature dissimilarities, while densely sampled data may enhance localisation precision but decrease performance, as overlapping satellite image patches increase the likelihood of incorrect retrievals

\section{Methodology}
\subsection{City-Scale Geo-Localisation Data Representation} 

We frame \ac{cvgl} as a graph comparison problem, similar to the technique demonstrated in \textit{SpaGBOL} \cite{spagbol}. 
Where \textit{SpaGBOL} established a lower bound on localisation precision by only applying graph nodes at road junctions, we incorporate orders of magnitude more nodes by placing \textit{secondary nodes} along existing edges, enhancing the density of data.
These graphs now have two classes of nodes, denoted \textit{primary} nodes $N$ - representing road junctions, and \textit{secondary} nodes $Q$ - captured along roads at varying intervals.
This significant increase in data density greatly increases the precision upper bound.
Figure \ref{fig:london_graph} shows a section of this graph representation of Manhattan.

\vspace{0.4em}

We represent each region in the dataset $i \in \{Manhattan, ...\}$ as a separate graph $G_i = (N, Q, E)$ with primary nodes $N_{i} = \{n_{1}, n_{2}, ..., n_N\}$, secondary nodes $Q_{i} = \{q_{1}, q_{2}, ..., q_Q\}$, edges $E_i = \{e_{1,2}, e_{1,3}, ..., e_E \}$. 
Edges $e_{a,b}$ represent roads connecting primary nodes ${a}$ and ${b}$.
Each node in both classes has attributes - $\{I_{sat}, I_{street}, L, \Psi, B \}$, containing a panoramic streetview image and a satellite image - both RGB: $I_{j} \in \mathbb{R}^{3{\times}W{\times}H}, j \in \{street, sat\}$,  location $L=\{\phi, \lambda\}$ consists of geographical latitude and longitude coordinates, $\Psi \in \mathbb{R}: \{-180\degree \leq \Psi \leq 180\degree\}$ is the north-aligned camera yaw, and $B=\{\beta_{1}, ..., \beta_{K}\}$ are north-aligned bearings to $K$ neighbouring nodes - where $\beta \in \mathbb{R}: \{-180\degree \leq \beta \leq 180\degree\}$.

\begin{figure}[t!]
  \centering
    \includegraphics[width=0.91\columnwidth]{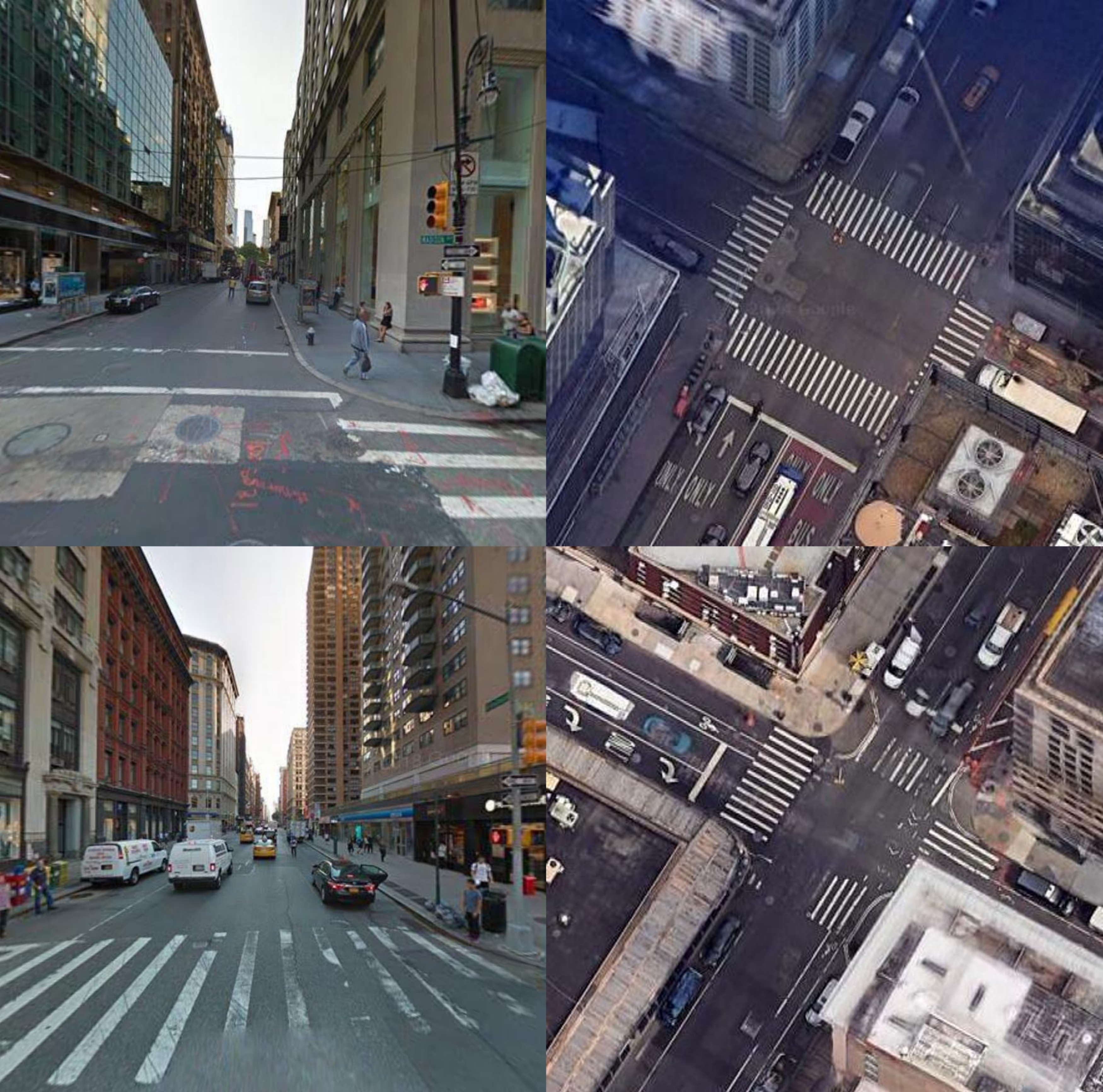}
    \caption{Example primary node (road junction) cross-view image pairs. Left-hand side shows $90\degree$ crops from panoramas and the right-hand side shows aerial images at zoom 20.}
    \label{fig:neighbourhood}
\end{figure}

\begin{figure*}[t!]
  \centering
    \includegraphics[width=\textwidth, height=300px]{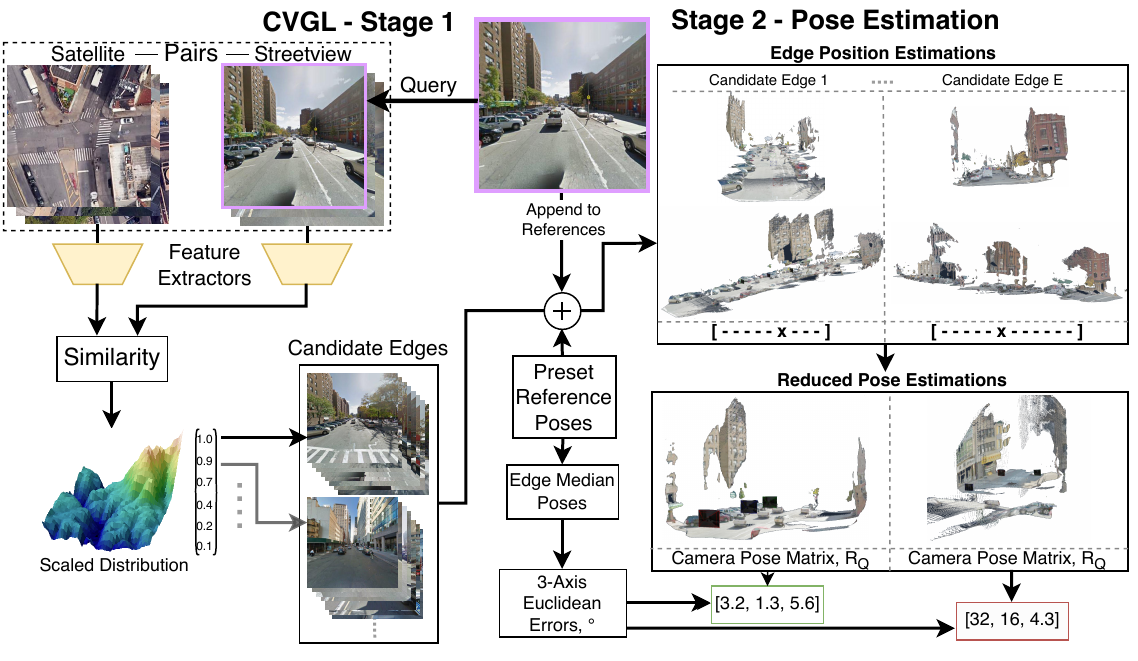}
    \caption{2-Stage system diagram. 
    Stage 1 retrieves scaled similarities of reference embeddings for the latest seen primary node, acquiring ordered candidate edges. 
    Stage 2 runs through edges consecutively until a threshold is met or completion. Position along an edge is estimated against all reference images, then estimating pose with the predicted adjacent two images.}
    \label{fig:system}
\end{figure*}

We limit the streetview image's ($I_{street}$) \ac{fov} to increase the technique's feasibility as a large proportion of existing vehicles possess monocular cameras. 
Cameras are assumed to be fixed to the vehicle in a forward-facing configuration. 
We experiment with \ac{fov}s, $\Theta \in \{70\degree, 90\degree, 120\degree\}$.

\subsection{PEnG Procedure}
Our proposed technique, \ac{paper_name}, operates in two stages, described in Figure \ref{fig:system}: initially estimating \textit{candidate primary nodes} with graph-based \ac{cvgl} (shown on the left-hand side) before performing \ac{rpe} relative to the secondary nodes present along each \textit{candidate edge} until a threshold is met, or all candidate edges have been processed.

The main purpose of the first stage is to reduce the number of reference images when performing relative pose estimation. 
This enables city-scale pose estimation as without it, pose estimation takes orders of magnitude longer. 

\vspace{0.4em}

\subsubsection{Graph-Based Cross-View Geo-Localisation}
We perform \ac{cvgl} following the standard procedure as used within previous works \cite{shore2023bevcv, 8578856, Shi2020WhereAI}.  
We implement a siamese-like network of CNN feature extractors, with no weight sharing, to produce similar embeddings $\eta_{t}$ from corresponding streetview-satellite image pairs. 
Creating a database of reference embeddings offline, querying this database for retrievals during online operation.

\vspace{-0.4em}
\begin{equation}
    \eta_t = \mathrm{CNN}\left(I_t | \omega_t\right), t \in \{street, sat\}
    \label{eq:feat_street}
\end{equation} 

In the first stage, \ac{cvgl} retrievals are only performed on primary nodes $N_{i}$ to provide efficient and accurate initial filtering.
Retrieved reference embeddings are ordered by descending similarity with the query, and are then min-max normalised to between 0 \& 1 giving a confidence score $c_i$ for each candidate node - concluding this stage.
Top candidate nodes, $C_k$, are passed to the second stage depending on the minimum confidence threshold $\theta_c$, and maximum number of candidates $k$. 
\begin{equation}
    c_i = \text{scale}\biggl(\frac{\eta^{query}_i \cdot \eta^{ref}}{\|\eta^{query}_i\| \|\eta^{ref}\|}, 0, 1\biggr)
    \label{eq:cvgl}
\end{equation}
\begin{equation}
    C_k = \{ c_i | c_i > \theta_c \text{ and } i<k\}
    \label{eq:cvgl_conf}
\end{equation}

\subsubsection{Pose Refinement}

\begin{figure}[!tbp]
    \includegraphics[width=\columnwidth]{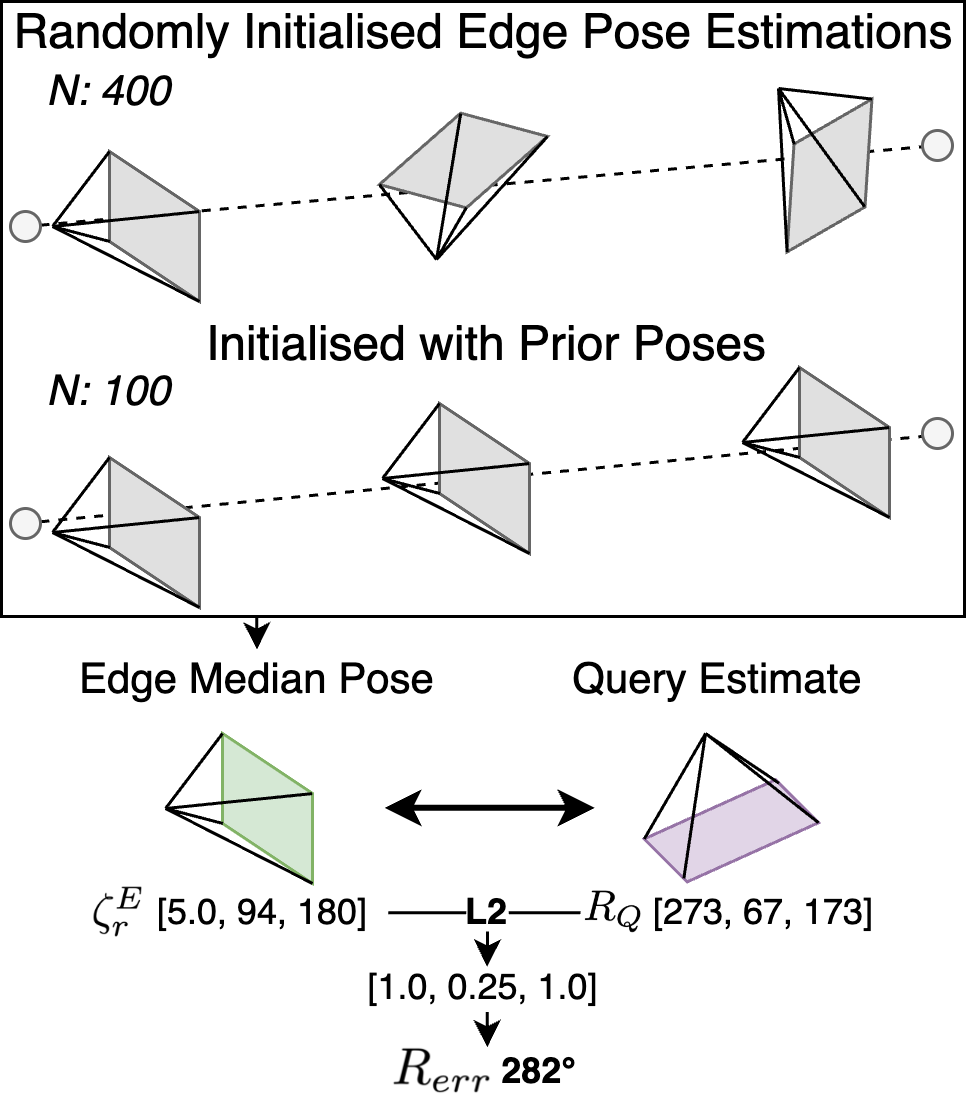}
    \caption{Pose estimates within each candidate edge are scored by their 3-axis euclidean distance with the mean rotational pose of the secondary nodes. This is possible due to the known orientations of edges within graph representations.}
    \label{fig:medianpose}
\end{figure}

For each candidate node, $c$, we select that candidate's connected edges, $E_{c} = \{e_{i,j} | i=c \text{ or } j=c\}$. We then filter these edges by matching the compass heading and the edge's yaw within the graph.
For every remaining candidate edge, we then perform \ac{rpe} in two stages: first estimating a coarse position of the query image along an edge before refining this relative to the two neighbouring reference secondary nodes. 
The calculation of median edge rotational pose is displayed in Figure \ref{fig:medianpose}.

\vspace{0.4em}

Inspired by \cite{leroy2024groundingimagematching3d}, we determine the relative pose of query images against each candidate edge's secondary node, before combining the poses across the entire edge.
For each image pair along an edge $I^1$ \& $I^2$, we determine the set of cross-image pixel correspondences. 
We then use a transformer-based network to predict 3D pointmaps, $X^{1,1}, X^{1,2}$, from 2D points $x^i$ between these images, expressed in the coordinate frame of $I^1$.
The pointmaps are then compared $X^{1,1} \longleftrightarrow X^{1,2}$, computing the relative poses with RANSAC \& PnP \cite{RANSAC} expressed in equations \ref{eq:pnp} and \ref{eq:pose}.

The objective of PnP is to minimise the reprojection error between the 3D points and their corresponding 2D image projections:
\begin{equation}
    x^i = K(RX^i + t)
    \label{eq:pnp}
\end{equation}
Where $x^i$ is the projected 2D point, $X^i$ is the 3D world point, $K$ is the estimated camera intrinsic matrix, $R$ \& $t$ are the rotation and translation matrices.
RANSAC randomly samples 4 points for PnP, optimising the objective to estimate $R$ and $t$. 

We compute the reprojection error as $e_i = ||x_i - K(RX_i + T)||$, rejecting outliers based on a predefined threshold $\epsilon$.
We then maximise the number of inliers $e_i \leq \epsilon$ to achieve the best pose estimate $(R^*, t^*)$: 
\begin{equation}
    (R^*, t^*) = \underset{R, t}{\text{argmax}} \, \sum_{i} \mathbf{1}(e_i \leq \epsilon)
    \label{eq:pose}
\end{equation}
where $(e_i \leq \epsilon)$ is the indicator function - equals $1$ if $e_i$ is less than or equal to a predefined threshold $\epsilon$, $0$ otherwise. \\

\textbf{Precomputation} - All reference poses, $P^r$, are estimated prior to system operation, calculating a median 3-DoF rotational matrix for each edge $\zeta_{r}^{E}$.
As this is a preprocessing step, a larger number of iterations are used compared to during inference.
These pre-determined poses then initialise optimisation processes during operation, reducing the required number of iterations - leading to lower operating times without effecting performance.

\vspace{0.4em}

\textbf{Operation} - Algorithm \ref{al:peng} is executed for each query image, until thresholds such as Maximum Rotational Error $\theta_{re}$ or No. Candidate Nodes $\theta_{n}$ are achieved. 
Rotational error $R_{err}$ is the 3-DoF summed euclidean distance between the query rotation $R_Q$ and the median edge rotation $\zeta_{r}^{E}$. This is calculated with an $[X, Y, Z]$ axis weighting of $[1, 0.25, 1]$ as roll has a smaller impact on performance. 
Where a query has multiple pose estimations and an L2 distance threshold has not been met, each pose is given a confidence score - rotational errors are summed and min-max scaled to between 0 \& 1.
Confidence scores from both stages are considered to determine a final pose estimation, calculated by scaling the relative poses to between the edge's ground truth limits.

\begin{algorithm}[t!]
\caption{PEnG Algorithm}
\label{al:peng}
\begin{algorithmic}[1]
    \Require 
    Graph $G=(N, Q, E)$,
    Reference Primary Node Database $\eta^{sat}_{N}$,
    Query and Reference images $I^{street}_{Q}$ $I^{sat}_{R}$,
    Thresholds $\theta_{x} \in \{\theta_{pe}, \theta_{n},...\}$, 
    Reference Poses $\zeta_{r}^{E}$

    \Ensure $R_Q$
    
    \State \textbf{Stage 1 - CVGL}
    \State $\eta^{street} = \text{CNN}(I^{street})$
    \State $S = \text{scale}\biggl(\Bigl(\sum_{k}\eta^{ref}_{ik}\eta^{query}_{k}), 0, 1\Bigr)$ \\

    \State \textbf{Stage 2 - Pose Estimation}
    \State $i = 0$
    \While {$\text{thres}(R_{err} \leq \theta_{x})$} 
    
        $E_{cand} = \text{filter}(N(S_i), \Psi)$

        $I^{pairs} = \textit{exhaustive}(E_{cand} + I^{street})$

        $t_{p} = \text{RPE}_{position}(I^{pairs})$

        $(R^i, t^i) = \text{RPE}_{pose}(I^{t_{p-1}}, I^{t_{p+1}})$
        
        $R_{err}^i = \text{sim}_{euc}(R^i, \overline{E_{cand}})$

        $i = i + 1$
    \EndWhile

    \State

    \State \Return Absolute Pose Estimations $R_Q$

\end{algorithmic}
\end{algorithm}

\vspace{0.4em}

\section{Results}

\begin{figure*}[!tbp]
    \vspace{0.5em}
    \includegraphics[width=\textwidth]{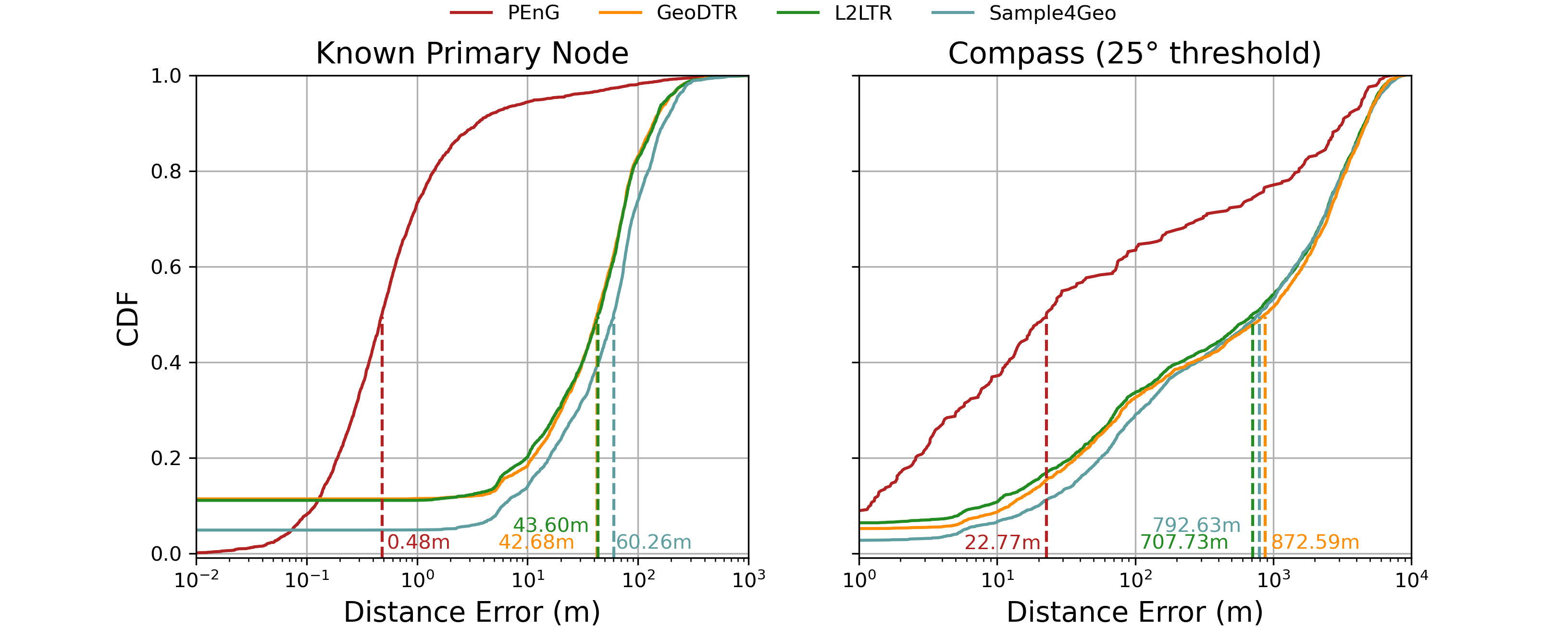}
    \caption{Cumulative Distribution Functions show the significant decrease in distance error achieved with \ac{paper_name}. 
    Previous works are non-zero at $x=0$ as there is $0$m error when they correctly retrieve the corresponding correct satellite image.}
    \label{fig:cdfs}
\end{figure*}

\subsection{Datasets}
The feature extractors for both \ac{paper_name} and previous works are trained with the CVUSA dataset \cite{cvusa}, cropping streetview images to various \ac{fov}s, portraying front-facing road-aligned monocular images. 
This dataset contains $35,532$ streetview-satellite training pairs and $8,884$ validation pairs. 
CVUSA satellite images have a resolution of $750 \times 750$ and streetview panoramas of $1232 \times 224$, both north-aligned.
We evaluate with the StreetLearn Manhattan dataset \cite{streetlearn}. 
Example image pairs are shown in Figure \ref{fig:neighbourhood}.
Manhattan is selected for evaluation as it qualifies as an urban canyon - an environment category that often experiences GNSS failure. 
The city's data are converted from unconnected images into a graph representation. 
This contains $53,289$ images, comprising $2,622$ primary nodes and $50,667$ secondary nodes. 
The graph covers approximately $31.6\text{km}^2$. 
Satellite images are north-aligned with a resolution of $0.20\text{metres} / \text{pixel}$ covering $50\text{m}^2$ (some images may have been captured from drones and other aerial image sources). 
Streetview images are yaw-aligned panoramas with a resolution of $1664 \times 832$. The median distance between the primary nodes is $116$m, and the median distance between adjacent secondary nodes is $9.83\text{m}$.
As both training and evaluation datasets contain camera yaw values at image capture, we are able to produce limited-FOV front-facing crops, emulating a monocular camera - our expected input for real-world CVGL application for autonomous vehicles.

\subsection{Implementation Details}
Image features are extracted with a ConvNext-T \cite{liu2022convnet} pre-trained on ImageNet-1K \cite{5206848}, producing 768-dimension embeddings.
When evaluating against SpaGBOL \cite{spagbol} we instead use their trained feature extractor - a combination of a ConvNext-T CNN with a GraphSage GNN, generating low-dimensional vector representations. We perform this second evaluation with randomly sampled depth-first walks from the graph.
We filter candidate edges by emulating a compass alongside the query, discarding incompatible graph edges. 
This is possible due to the graph representation - with known orientations between the primary node and it's connected edges. 
All existing CVGL baselines are also augmented with this compass filtering technique to ensure a balanced assessment.

\vspace{0.4em}

We use a median pose error threshold of $3\degree$, halting execution if a match is found with a weighted euclidean distance below this.
In the rare case that all edge pose estimates have an error larger than this threshold, the estimate with lowest error is selected.
The feature extractor is trained with FOVs $\in \{70\degree, 90\degree, 120\degree\}$ for 50 epochs using an AdamW optimiser with an initial learning rate of $1e-4$ and a ReduceLROnPlateau scheduler. 
The preset poses stored for reference points are calculated offline with a learning rate of $0.1$ and $400$ iterations, which are refined when online with a learning rate of $0.1$ and $100$ iterations.

\vspace{0.4em}

\subsection{Ablation Study}
To verify the contribution of each constituent in the proposed system, we display an ablation study in Table \ref{tab:ablation}. 
CVGL shows the performance of the simple ConvNeXt-T feature extractor, evaluated in the same method as previous works - filtering by primary nodes initially to reduce the reference set.
\textit{1 Pose} performs pose estimation against an entire edge's reference images, determining a relative 2-DoF pose between primary nodes. 
\textit{2 Pose} follows \textit{1 Pose} with a refined pose estimation relative to the 2 adjacent reference secondary nodes, determined in the first pose estimation step - this enables a high precision final estimate.
\textit{Pose Priors} is the addition of estimating the pose of all secondary nodes prior to querying, increasing the accuracy of reference poses and offloading a portion of computation to an offline stage. 

\begin{table}[tbp!]
\centering
\resizebox{\columnwidth}{!}{
\begin{tabular}{c|cccc}
Config & Med (m) & Top-1m & Top-5m & Top-25m \\ \hline
CVGL & 961 & 6.06 & 6.94 & 9.27 \\
1 Pose & 32.12 & 7.02 & 25.45 & 45.12 \\
2 Pose & 28.94 & 7.31 & 26.41 & 47.91 \\ 
Pose Priors & \medninety{} & \toponeninety{} & \topfiveninety{} & \toptwentyfiveninety{} \\
\end{tabular}
}
\caption{Successive ablation of \ac{paper_name} stages to demonstrate the contribution of each, with $90\degree$ horizontal FOV.}
\label{tab:ablation}

\end{table}

\vspace{0.4em}

The ablation shows the vast decrease in median distance error achieved by combining these two localisation techniques, the median error decreases by an order of magnitude.
Having a pose refinement stage after the initial position estimation further decreases median error by $\approx3$m.
Finally, estimating reference poses prior to operation increased accuracy relatively by $\approx10$\%.

\vspace{0.4em}

\subsection{Evaluation}
We evaluate with distance-based Top-K recall accuracy, displaying euclidean distance errors in \ac{cdf} plots - displayed in Figures \ref{fig:cdfs}. 
Table \ref{tab:sota} shows discretised metrics for these functions, defining estimates as successful if they are within K-metres of the ground truth. 
We evaluate how \ac{paper_name} performs with images of varying \ac{fov}, with higher-\ac{fov} cameras tending to be more expensive but able to capture more information.
All comparisons follow the 2-stage process: first predicting the closest primary node, then estimating the closest position within the reduced subset of connected secondary nodes. 
To demonstrate the generality of the \ac{paper_name} approach we present results with both a traditional retrieval first stage, \ac{paper_name}, and a graph-based first stage, \ac{paper_name}*.

\begin{table}[]
\centering
\resizebox{\columnwidth}{!}{%
\begin{tabular}{c|cccc}
Model & \multicolumn{1}{c|}{Med (m)} & Top-1m & Top-5m & Top-25m \\ \hline
\multicolumn{1}{r|}{FOV} & \multicolumn{4}{c}{\textbf{$70\degree$}} \\ \hline
L2LTR \cite{L2LTR} & \multicolumn{1}{c|}{826} & 6.48 & 7.55 & 10.03 \\
GeoDTR+ \cite{GeoDTR+} & \multicolumn{1}{c|}{903} & 5.19 & 6.03 & 8.47 \\
Sample4Geo \cite{sample4geo} & \multicolumn{1}{c|}{897} & 6.79 & 7.78 & 10.41 \\
\textbf{PEnG} & \multicolumn{1}{c|}{\textbf{26.82}} & \textbf{7.25} & \textbf{27.43} & \textbf{49.01} \\ \hline
SpaGBOL \cite{spagbol} & \multicolumn{1}{c|}{634} & 6.37 & 7.70 & 10.56 \\
\textbf{PEnG*} & \multicolumn{1}{c|}{\textit{29.31}} & \textit{6.86} & \textit{26.16} & \textit{47.75} \\ \hline
\multicolumn{1}{r|}{FOV} & \multicolumn{4}{c}{\textbf{$90\degree$}} \\ \hline
L2LTR \cite{L2LTR} & \multicolumn{1}{c|}{750} & 6.64 & 8.01 & 10.64 \\
GeoDTR+ \cite{GeoDTR+} & \multicolumn{1}{c|}{854} & 6.06 & 7.25 & 9.80 \\
Sample4Geo \cite{sample4geo} & \multicolumn{1}{c|}{734} & 8.35 & 9.31 & 12.43 \\
\textbf{PEnG} & \multicolumn{1}{c|}{\textbf{\medninety{}}} & \textbf{\toponeninety{}} & \textbf{\topfiveninety{}} & \textbf{\toptwentyfiveninety{}} \\ \hline
SpaGBOL \cite{spagbol} & \multicolumn{1}{c|}{529} & 6.33 & 7.25 & 9.69 \\
\textbf{PEnG*} & \multicolumn{1}{c|}{\textit{34.91}} & \textit{7.17} & \textit{25.10} & \textit{47.29} \\ \hline
\multicolumn{1}{r|}{FOV} & \multicolumn{4}{c}{\textbf{$120\degree$}} \\ \hline
L2LTR \cite{L2LTR} & \multicolumn{1}{c|}{732} & 7.82 & 9.19 & 12.05 \\
GeoDTR+ \cite{GeoDTR+} & \multicolumn{1}{c|}{893} & 6.75 & 7.63 & 10.60 \\
Sample4Geo \cite{sample4geo} & \multicolumn{1}{c|}{703} & \textbf{9.50} & 10.68 & 14.42 \\
\textbf{PEnG} & \multicolumn{1}{c|}{\textbf{37.72}} & 4.04 & \textbf{21.21} & \textbf{44.93} \\ \hline
SpaGBOL \cite{spagbol} & \multicolumn{1}{c|}{501} & \textit{6.90} & 7.86 & 10.14 \\
\textbf{PEnG*} & \multicolumn{1}{c|}{\textit{45.46}} & 3.36 & \textit{22.04} & \textit{44.24} \\ \hline
\end{tabular}%
}
\caption{Localisation precision comparison to previous works with a stage 1 scoring 0.9 threshold. Best image pair method displayed in \textbf{bold}, best graph-based method shown in \textit{italic}.}
\label{tab:sota}


\end{table}

\vspace{0.4em}

To increase fairness in comparison against traditional single-stage \ac{cvgl} works, we augment these baselines with a secondary refinement stage where the same technique is run again, but only required to match against the ground-truth satellite images of the corresponding secondary nodes. 
In a real-world use case this is infeasible, as the reference set cannot contain precisely geographically aligned ground truth satellite images. 
However, it serves to provide a stronger baseline for comparison.

\vspace{0.4em}

The evaluation shows that our proposal achieves significant improvements over current \ac{sota}. With $90\degree$ images, we achieve a \meddecrease{}\% reduction in median error, and an approximate \topfiveincrease{}\% increase in Top-5m accuracy. 
We note that using $90\degree$ \ac{fov} images achieves a relative decrease in the median error of $\approx4$m compared to $70\degree$. 
This is due to the increase in information available to each stage.
However, further increasing the \ac{fov} to $120\degree$ yields a decrease in localisation precision.
This may be caused by the input image dimensionality limitation of our model - due to the backbone pre-training, the maximum image resolution for the system is $512 \times 384$, placing an upper bound on how much information can pass through the system.
Another hindrance is experienced from extracting perspective images from a $360\degree$ panorama.
When increasing the horizontal \ac{fov} beyond $90\degree$, these crops begin to display visibly distortion.

\vspace{0.4em}

Within the discretised Top-Km metrics, \ac{paper_name} performs slightly worse than previous works where $K < 5$ due to the inherent zero error bias in existing CVGL works.
As $K$ reaches 25m, performance is significantly higher across \ac{fovs}.
As precisely centred ground-truth corresponding satellite images are known for each query streetview image in \ac{cvgl}, they tend to perform unrealistically well with these Top-K metrics.
This peculiarity of previous evaluation protocols is visible in Figure \ref{fig:cdfs} where at $x=0$, previous works start from a non-zero values.

\section{Conclusion \& Future Work}
We successfully propose and demonstrate the utility of combining graph-representations, \ac{cvgl}, and relative pose estimation techniques.
This ensemble is proven to be a viable strategy for progressing \ac{cvgl} within a large city-scale environment towards practicality, reducing median distance errors from hundreds of metres down to often centimetre level accuracy.
\ac{paper_name} achieves \ac{sota} localisation precision when evaluated within the Manhattan region of $36.1km^2$, reducing the median error from Sample4Geo's previous best of 734m down to \medninety{}m when operating with $90\degree$ FOV.
In our ablation studies, we thoroughly demonstrate the significance of each portion of the 2-stage architecture, validating that the combination results in the maximum precision possible for \ac{paper_name}.
We release code for converting the StreetLearn dataset into the graph representation outlined above, along with \ac{paper_name} technique's code and corresponding pretrained weights, enabling future works to build upon the technique and further evaluate this ensemble.

\subsection{Future Work}
Several aspects of this work will be the target for optimisation in order to further progress the field towards real-world application.
Due to the vast disparity in viewpoint within \ac{cvgl}, performance from the first stage limits the potential precision achieved in the second stage. A more probabilistic fusion technique could mitigate this.
Furthermore, the second stage of \ac{paper_name}, \ac{rpe}, can be computationally costly compared to the first stage. There is a trade-off between accuracy and complexity, based on the number of iterations performed with RANSAC+PnP. Future work could explore sequential extensions of the technique, introducing temporal priors into the position estimation, to further filter the reference set and reduce the number of iterations required.

\section{Acknowledgements}
This work was partially funded by the EPSRC under grant agreement EP/S035761/1 and FlexBot - InnovateUK project 10067785.

\newpage 

{\small
\bibliography{egbib}
}

\end{document}